# MechAgents: Large language model multi-agent collaborations can solve mechanics problems, generate new data, and integrate knowledge

Bo Ni[1], Markus J. Buehler[1,2,3]*

[1] Laboratory for Atomistic and Molecular Mechanics (LAMM), Massachusetts Institute of Technology, 77 Massachusetts Ave., Cambridge, MA 02139, USA

[2] Center for Computational Science and Engineering, Schwarzman College of Computing, Massachusetts Institute of Technology, 77 Massachusetts Ave., Cambridge, MA 02139, USA

[3] Lead contact

*Correspondence: mbuehler@MIT.EDU

**ABSTRACT**:

Solving mechanics problems using numerical methods requires comprehensive intelligent capability of retrieving relevant knowledge and theory, constructing and executing codes, analyzing the results, a task that has thus far mainly been reserved for humans. While emerging AI methods can provide effective approaches to solve end-to-end problems, for instance via the use of deep surrogate models or various data analytics strategies, they often lack physical intuition since knowledge is baked into the parametric complement through training, offering less flexibility when it comes to incorporating mathematical or physical insights. By leveraging diverse capabilities of multiple dynamically interacting large language models (LLMs), we can overcome the limitations of conventional approaches and develop a new class of physics-inspired generative machine learning platform, here referred to as MechAgents. A set of AI agents can solve mechanics tasks, here demonstrated for elasticity problems, via autonomous collaborations. A two-agent team can effectively write, execute and self-correct code, in order to apply finite element methods to solve classical elasticity problems in various flavors (different boundary conditions, domain geometries, meshes, small/finite deformation and linear/hyper-elastic constitutive laws, and others). For more complex tasks, we construct a larger group of agents with enhanced division of labor among planning, formulating, coding, executing and criticizing the process and results. The agents mutually correct each other to improve the overall team-work performance in understanding, formulating and validating the solution. Our framework shows the potential of synergizing the intelligence of language models, the reliability of physics-based modeling, and the dynamic collaborations among diverse agents, opening novel avenues for automation of solving engineering problems.

## 1. Introduction

Solving mechanics problems using numerical methods,[1–6] such as finite element methods (FEM),[7,8] has garnered broad applications in various engineering applications, ranging from elucidating mechanisms and understanding,[9,10] deformation/failure prediction,[11–13] to structure/material design.[14,15] Despite their popularity, these modeling techniques often require deep technical expertise and experiences to successfully perform such simulations in a meaningful way.[16–18] For example, to solve an elasticity problem using FEM, it demands not only knowledge of elasticity theory[19] and FEM[8] but also a capability of programming, debugging and postprocessing within certain simulation environments.[20–23] Therefore, so far it has mainly been reserved for human experts to practice. An area of intense interest in mechanics and related field has been the development of machine learning (ML) based strategies, and it has been shown as effective tools to provide effective approaches to solve complex end-to-end problems such as predicting mechanical properties of composites [24,25], stress and strain fields [26–29] or [30] conducting inverse design tasks[31–34]. This is often done via the use of surrogate models or structural deep learning methods such as attention-models or graph neural networks; however, in such models their knowledge is baked into the architecture during the training process, offering less flexibility when it comes to generating, collecting and incorporating new data or physical insights. Approaches such as active learning or Bayesian strategies (e.g. using Gaussian Process Modeling) provide alternative methods[35–38], but are typically programmed *a priori*.

Recently, large language models[39–43] (LLMs) based on attention mechanisms and transformer architectures (and various flavors thereof)[44,45] have been emerging as novel tools for various applications and fields, including int eh context of mechanics and related areas[46–49]. For example, the specific class of Generative Pretrained Transformer (GPT) models, such as the ChatGPT[50] and GPT-4[40] models reported by OpenAI, have demonstrated unprecedented capability in mastering human language, coding, logic, reasoning, and related natural language tasks, including leading complex conversations.[51,52] Recent studies have demonstrated that such complex LLMs excel in programming numerical algorithms or debugging code errors in various coding languages, such as Python, MATLAB, Julia, C and C++.[52–54] At the same time, based on their conversational capabilities, LLMs have been used to power conversable AI agents to go from AI-human conversation to AI-AI or AI-tools interactions for autonomy.[46,55,56] For instance, given a statement of a specific task, AI agents can attempt to break complex problem statements into subtasks and use tools, including data retrieval from the internet, to solve them step-by-step through automatic iterations.[57,58] For materials applications, agent-based modeling was first suggested in [48], building on earlier successes of using attention-based machine learning models to capture complex mechanics phenomena such as field predictions, fracture or dynamical phenomena, included in a multi-modal context[28,59–63]. Combining these advances, here we investigate whether, and how, we can integrate the language capabilities, both in natural and coding languages, of LLM powered agents with the numerical simulation tools to solve mechanics problems with reduced or little human intervention, featuring a broad set of tasks that involve problem identification, code development and debugging, plotting results and analysis, and providing interactive feedback with the human user.[42,28,52–54] To achieve this goal, we propose the use of multi-agent[55] frameworks. This proposes a new strategic direction by which scientific AI can support sold mechanics research and even generate new data based on known mathematical/numerical models (e.g., FEM) and sociological concepts (e.g., division of labor).

The plan of this paper is as follows. After an introduction into the strategic approach, we demonstrate the method in the solution of elasticity problems by creating a set of conversable agents with comprehensive or specialized roles by profiling and organizing them into collaborative teams to apply knowledge of elasticity and FEM to solve problems with little human intervention (**Figure 1**). Through self-correction, a two-agent team (**Figure 1b**) is shown to solve well stated elasticity problems with different boundary conditions, domain geometries and FEM meshes, small/finite deformation with linear/hyper-elastic constitutive laws, and post-processing. By introducing division of labor and mutual corrections, a group of agents tasked respectively with plan development, problem formulating, code writing, program executing and result criticizing can collaborate autonomously via conversation (**Figure 1c**) to solve significantly more challenging problems at which a two-agent team fails. Our multi-agent modeling frameworks demonstrate the promising possibility of going beyond the human-only paradigm for solving mechanics problems using available knowledge and reliable tools, thus opening novel avenue for automation of problem solving in engineering and the preparation of the coming era of human-AI teaming up in engineering research and innovation. We conclude the paper with a discussion of the results and a broader elaboration on the limitations and potentials of agent strategies.

## 2. Results and Discussion

To explore the potential of organized multi-agent modeling frameworks for solving mechanics problems we present a series of computational experiments to demonstrate the benefits of self-correction and mutual corrections via agents' teaming strategies and collaborative approach. As shown in **Figure 1a**, we power the agents through a state-of-the-art general purpose large language model, GPT-4,[40] via the OpenAI API.[64] Each agent has its own profile for initialization and they are organized into teams of different structures.[63] Depending on their profiles, they can explicitly communicate with each other by sharing information, or with human using language; and some of the agents are given access to a simulation environment in which they can execute simulation code.

We demonstrate the performances of agent teams by assigning tasks of solving classical elasticity problems using FEM via the python package, FEniCS.[65] Further details can be found in the **Materials and Methods** section. We start with a simple two-agent team and then extend to a multi-agent group with detailed division of labor, studying and comparing their performances in problem solving and error correcting for elasticity problems.

## 2.1 A two-agent team with self-correction

We begin with a basic two-agent team to demonstrate the benefits of self-correction via continuous conversation in solving mechanics problems. As shown in **Figure 1b**, the team consists of a human user proxy agent and an assistant agent. The former can pass the task assignment from human user to the assistant agent, execute the code generated by the assistant and return the outcome of the execution; the latter is in charge of formulating the plan for solving the assignment, writing the simulation code, responding to the outcome of the execution of the code and signaling to terminate the conversation when it thinks the solution is satisfactory. In this team structure, we intentionally let the assistant agent play the key role in handling comprehensive tasks, including understanding the assignment, formulating an effective plan, writing and debugging the solution code and evaluating the results. To test its capability, in one conversation, we start with a basic task and then assign further demands on different aspects of elasticity problems until the length of the conversation reaches the token limit of the LLM. We provide specific instruction only when the assistant agent fails to realize its own failure or error.

*Conversation 1: Linear elasticity problems*

In this conversation, we focus on linear elasticity problems. To start the first round, the following task is assigned through the user proxy agent:

> **user_proxy (to assistant):**
>
> > A 1m-by-1m elastic plate has Young's modulus of 1GPa and Poisson ratio of 0.3. It has zero displacement on the left edge and 0.1m displacement along x direction on the right edge. Solve for the displacement using FENICS and store the displacement result in a PNG file.

In response, the assistant agent states the plan of using FEniCS to solve this problem, generates the python script and explain its function. Benefited from the knowledge of GPT-4, this initial script correctly formulates the weak form of linear elasticity problem but forgets to define the prescribed material properties (*i.e.*, Young's modulus and Poisson's ratio). However, when the proxy agent sends back the error information generated by executing the script to the assistant agent, it successfully identifies the cause of the error by stating the following, before proceeding to revising the script:

> *I apologize for the oversight. The variables `lambda_` and `mu` are indeed not defined in the script. These are Lame's constants and they can be calculated from Young's modulus (E) and Poisson's ratio (nu) as follows:*
>
> ```python
> E = 1e9  # Young's modulus in Pa
> nu = 0.3  # Poisson's ratio
>
> mu = E / (2 * (1 + nu))  # Shear modulus (Lame's first parameter)
> lambda_ = E * nu / ((1 + nu) * (1 - 2 * nu))  # Lame's second parameter
> ```

This self-correction via continuous conversation between the two agents enable the team to handle the initial failure and improve the script continuously. After two rounds of such debugging, the team renders a correct script and executes it to properly solve the assigned problem. The typical flow for self-correcting observed between the two agents is summarized in **Figure 2a**. Note that, the logic of this flow is not hardwired or defined *a priori* but emerges naturally as a continuous conversation. The displacement result generated by the two-agent team, without human intervention, is shown **Figure 2b**. The complete conversation and modeling scripts can be found in **Table S1.1** in **Section S1.1** of the **Supplementary Materials** section.

To test the capability of this two-agent team in handling different aspects of linear elasticity problems, we continue the conversation by assigning various more complex tasks. As listed in **Table 1** Column 2, in the following rounds of conversation, we ask the team to resolve the problem considering changing the boundary condition from tension to shear, refining the finite element mesh (Round 2), adding a circular hole in the geometry (Round 3) and extracting the shear stress component (Rounds 4 and 5) step by step. The main steps taken by the assistant agent are summarized in Column 3 and the results are inspected by the human authors in Column 4 in **Table 1**. The corresponding output files generated by the agent team are shown in **Figure 2c-f.** For most of the demands, the team is able to make necessary modifications to the simulation script, self-correcting when facing errors after executing, and solve the new assignments correctly (Round 1-3).

At the same time, there exists situations where the generated script runs smoothly without errors but the assignment is not necessarily fulfilled, which make it harder for the assistant agent to do self-correction. For example, in Round 4, when asked to extract the shear stress component, $\sigma_{xy}$, the assistant agent mistakenly extracts von Mises stress (**Figure 2e**) in the script. When the execution of the script is successful, the assistant agent naively concludes the task without realizing its mistake. This kind of conceptual error may require external intervention to point the assistant agent towards the right direction. For instance, in Round 5 (**Table 2,** last row), when instruction from human intervention is provided, the assistant agent is able to correct this mistake in one shot and solve the assignment (**Figure 2f**). Note that, this external critics or evaluation may also be provided by other agents specialized in such tasks, thus reducing human intervention and improving automation in handling more challenging tasks in mechanics problems. This possibility and the corresponding multi-agent model will be discussed in the next subsection. The full conversation can be found in **Table S2-S5** in **Section S1.1** of the **SM**.

*Conversation 2: From linear elasticity to hyperelasticity*

To go beyond linear elasticity cases, in Conversation 2, we task the two-agent team with uniaxial tension problems with linear and hyper-elastic constitutive laws. As shown in **Table 2**, we start with a linear elastic material and then ask to change into a compressible Neo-Hookean material[19]. Similar to Conversation 1, the two-agent team may not solve the problems at the first attempts, but is able to self-correct based on the error information received and finally solve the two problems without any human instructions in the middle. The final scripts for the two assignments (Round 1 and 2 in **Table 2**) are shown in **Table 3**, where linear solver in FEniCS is used to solve the linear elastic case with small deformation assumption for strain (left column in **Table 3**) while nonlinear solver is used for the hyperelastic case considering finite deformation (right column in **Table 3**) respectively. The displacement results generated for the two cases are different as shown in **Figure 3**. The full conversation can be found in **Table S6-S7** in **Section S1.2** of the **Supplementary Material**.

Combining the simulation experiments above, we have demonstrated that the two-agent team is capable of self-correction via conversation flow and can solve classical elasticity problems of various boundary conditions, geometries and FEM meshes, small/finite deformation and linear/hyperelastic constitutive laws with little human guidance during the process.

**2.2 A multi-agent team using division of labor**

The conversation flow in the two-agent team enables the assistant agent to practice self-correction based on the outcomes of code execution, or other errors that may emerge. However, sometimes this self-correction is not sufficient in handling conceptual errors without explicit error messages (e.g. from running the code). This is similar to one human mind may fall short in realizing its own mistakes without external review or inputs. At the same time, the assistant agent is burdened with significant work of quite different nature, including making plans, formulating theories, writing and debugging the scripts and analyzing the results. This may further increase the challenge for a single agent to handle all of these effectively. To overcome these challenges, we adopt inspirations from human organizations and apply division of labor among a larger group of multiple (>2) agents (each with a certain assigned set of tasks, skills and/or profile) and explore its capabilities in solving mechanics problems.

As shown in **Figure 1c,** we design agents that play the roles of the administrator (admin), the planner, the scientist, the engineer, the executor, the critic and the group-chat manager, respectively, and organize them into a research group via autonomous, interactive and dynamic group chatting. The different role of each agent is defined via agent profiling (giving specific instructions via a system message). As listed in **Table 4**, different profiles (Column 3) for agent roles (Column 2) are written as initial prompts to influence the LLMs' behaviors during the chat. For example, the administrator is defined as a human proxy agent who assigns the task and can provide insights or choose to be silent when being asked; The planner is an agent who is tasked to develop a step-by-step plan and suggest specific subtasks for all other working agents to collaboratively solve the assignment. The scientist is in charge of formulating the mechanics problem using FEM and the engineer writes and debugs the code for implementation. The executor can run the script with the access to a simulation environment and return the outcome information back the group chat; the critic is specialized in providing critical evaluations for other non-human proxy agents and the whole solving process. To organize these working agents, we construct a group chat manager agent; this agent coordinates a dynamic conversation among the group by performing the following steps repeatedly:

- Choosing a speaker based on the context and the agent profiles
- Collecting the inputs from the selected speaker, and
- Broadcasting the message to the whole group.

To test the capability of this multi-agent group with division of labor in solving elasticity problems autonomously, we use the admin agent only to assign the tasks. While in principle it is possible to gather human input at various stages of the problem solving process, when being asked during the group chat, we choose to skip and provide no additional input from human. The performance of this multi-agent group is now discussed in comparison with that of the two-agent team above in the following case studies.

*Group chat 1: Linear elasticity problems*

As a reference case, we test this multi-agent group with the same assignments in Conversation 1 from the previous subsection. Organized by the group chat manager agent, the group chat enables close interactions between different agents and solve the assigned elastic problems effectively. As summarized **Figure 4a**, after given the task, the chat often starts with the planner proposing a step-by-step plan with tasks for individual agents. Then it is often the engineer who starts to the write FEniCS script and sends it to the executor. Based on the outcome of the execution, the engineer may revise the code several times until there is no error. Next, the scientist may comment on the script and results. At last, the critic provides critical evaluation about the results as well as whole process. Besides this seemly linear chat flow (solid arrows in **Figure 4a**) based on the division of labor, there exist rich interactions between the agents (dash arrows in **Figure 4a**) dynamically. For example, in the early stage, the critic may be chosen to speak after the planner and provides opinions on revising the plan; in the later stage, the critic may also jump in to provide suggestions on the script before or after its execution. This flexible topology of group chatting is not possible in the previous two-agent team and is expected to help the agents focus on specific tasks as well as mutually correct each other.

Using this multi-agent group with division of labor and dynamic chatting topology, the assignments in Conversation 1 are solved correctly without human guidance during the process. The generated files on displacement and shear stress are shown in **Figure 4b-e**. Note that, in the task of extracting shear stress component (Row 5, Column 2 in **Table 1**), the multi-agent group succeeds in its first attempt requiring no human input, thus surpassing the two-agent team who made this mistake without self-realization. The complete group chat records can be found in **Table S9-S12** in **Section S2.1** of the **Supplementary Material**.

*Group chat 2: On the effect of mutual correction*

To further test the capability of the multi-agent group, we modify the statement of the elasticity problem as the following.

> *A 2D plate occupies 1m-by-1m domain. It is made of copper and has a circular hole of radius 0.2m in the middle. It has zero displacement on the left edge and 0.02m displacement along x direction on the right edge. The top and bottom edges are free to move. Please use FENICS to solve the displacement, store the result in a PNG file and figure out the total force on the right edge.*

In this task statement, we avoid explicitly mentioning elasticity theory, identify the material name instead of prescribing the elastic properties, and ask for the traction force, which raises the difficulty for agents in formulating the problem, retrieving related material information and post-processing the results.

When tested with this task, the two-agent team constructed in the previous subsection produces a bug-free script (**Figure 5a**) and is satisfied with its outcome (**Figure 5c**) by stating the following.

> **assistant_OA (to user_proxy_OA):**
>
> *The code has been executed successfully and the total force on the right edge of the plate is 0.0. This result is expected because the displacement is only in the x-direction and the top and bottom edges are free to move, so there is no force acting on the right edge.*
>
> *The displacement plot should be saved as 'displacement.png' in the same directory where you ran the script. You can open this file to visualize the displacement of the plate.*
>
> *Please let me know if you need further assistance or if there's anything else you'd like to do. If everything is clear, I'll end this task.*

However, after inspection, we find the team fails the task with the following major mistakes: a) the circular hole is missing in the defined geometry; b) the assistant agent fails to formulate the correct weak form for an elasticity problem; c) the necessary material properties are not collected; d) the formulation for calculating traction is wrong (see the code blocks in the red dash rectangles **Figure 5a**). More importantly, as quoted above, when receiving the zero-traction force result, the assistant agent still thinks the result is right by providing some self-conflicting arguments (highlighted in red above) for it. The failure of the two-agent team indicates the updated task is more challenging and that the self-correction capacity in the two-agent team is not sufficient to handle it. The full conversation can be found in Table S8 in **Section S1.3** of the **SM**.

Next, we assign the same task to the multi-agent group with division of labor and the inspect the chat flow and the results. The final script generated (**Figure 5b**) by the group completes the task correctly and suffers no errors observed in the two-agent team case (the corresponding code blocks are indicated by the blue rectangles in **Figure 5b**). The correct displacement field file is shown in **Figure 5d.**

A careful inspection of the group chat flow indicates that the improvements achieved by the multi-agent team can be attributed to the division of labor and mutual corrections between agents. Some examples of correcting are listed here. For instance, after receiving suggestions from the critic, the planner revises the plan and clearly states the governing equation (highlighted in red) of the assign task as response.

> **Planner (to chat_manager):**
>
> *Revised Plan:*
> *...*
>
> *4. Engineer: Solve the PDE using FENICS. This will give the displacement of the plate. The elasticity equation to be solved is given by: div(sigma(u)) = 0, where sigma is the stress tensor and u is the displacement.*
> *...*

Correspondingly, in the first draft of the FEniCS script the engineer writes, the weak form for this governing equation is correctly formulated, which is a clear improvement compared with the two-agent team case. To calculate stress, the engineer also successfully retrieves the needed material properties for copper after analyzing the error information from executing the initial script.

> **Engineer (to chat_manager):**
>
> *I apologize for the oversight. The material properties for copper, including the Lame's first parameter (lambda_) and the shear modulus (mu), were not defined in the code. Let's correct this. The values for copper are lambda_=76.9 GPa and mu=48.3 GPa. Here is the corrected code:*
>
> *...*

At this stage, the script generated by the engineer still shares some major shortcomings with the one generated by the two-agent team in **Figure 5a**, including missing the circular hole in the geometry definition and the way of calculating traction on the right edge is wrong. The comments from the critic play the key role in identifying and solving these two errors. For example, even without executing the code, the engineer is able to include the circular hole part after the following comments (highlighted in red) from the critic.

> **Critic (to chat_manager):**
>
> *The revised plan and code seem to be well-structured and cover all the necessary steps to solve the problem. However, the code still lacks the implementation of the circular hole in the middle of the plate. This is a crucial part of the problem statement and should be included in the code.*
>
> *...*
>
> *Please revise the plan and the code to include these points.*

Similar to the case with the two-agent team, the bug-free script generated by the engineer initially reports a zero-traction result, which makes no sense for uniaxial tension test of a copper plate. In two-agent team case, the assistant agent fails to identify or criticize this unphysical result. Surprisingly, in the multi-agent group the critic is able to not only catch this abnormal clue but also provides comprehensive analysis for identifying the potential causes of it (see the quote in the following).

> **Critic (to chat_manager):**
>
> *The code execution was successful, but the total force on the right edge is reported as 0.0, which seems incorrect. This could be due to the fact that the stress tensor `sigma(u)` is not correctly defined or the force calculation is not correct.*
>
> *The stress tensor for linear elasticity is usually defined as `sigma(u) = lambda_*nabla_div(u)*Identity(d) + 2*mu*epsilon(u)`, where `lambda_` and `mu` are Lame's first parameter and the shear modulus, respectively, `nabla_div(u)` is the divergence of the displacement, `Identity(d)` is the identity tensor, and `epsilon(u)` is the strain tensor.*
>
> *The force calculation `force = dot(dot(sigma(u), n), Constant((1, 0)))*ds(2)` is supposed to compute the force on the right edge by integrating the stress over the area of the right edge. However, `ds(2)` might not correctly represent the right edge.*
>
> *Please double-check the stress tensor definition and the force calculation.*

With the help of the critic, the engineer is able to identify the cause (which is the wrong definition of the right edge and is included in the proposed causes by the critic) and revise the code to complete the task.

> **Engineer (to chat_manager):**

> *I apologize for the oversight. The force calculation is incorrect because the integration over the right edge is not correctly defined. The `ds(2)` in the force calculation is supposed to represent the right edge, but it seems that the right edge is not correctly marked as 2. Let's correct this. Here is the corrected code:*
>
> *...*

At last, provided with the results from the revised script, the critic is able to identify the success of the task and summarize the whole process clearly as the following and the complete group chat can be found in **Table S13** in **Section S2.2** of the **SM**.

> **Critic (to chat_manager):**
>
> *The code execution was successful and the total force on the right edge is reported as approximately 2052608504.74 Newtons. This seems to be a reasonable result for the given problem. The code correctly defines the problem, solves it using FEniCS, and computes the total force on the right edge. The code also correctly handles the boundary conditions and the circular hole in the middle of the plate. The displacement field is saved to a PNG file as required. The units for the displacement and the force are correctly specified as meters and Newtons, respectively. Overall, the code appears to be correct and complete.*

It should be noted that, all the agents, in either the two-agent team or the multi-agent group, are based on the same large language model, GPT-4. Thus, the different performances of the two organizations can likely be attributed to the effect of the structure and interactions between the agents that offer added synergies and emergence of capabilities. By solving the updated task that requires identifying problem type, retrieving material information, and handling complex post-processing, the multi-agent framework has demonstrated more robust performance via division of labor and mutual corrections. By focusing a specific task, the agent produces more robust performance in reasoning, planning and coding; by introducing a dynamic group chatting topology, agents, especially the critic agent, can provide valuable feedbacks to other working agents. This mutual correction performs better than the self-correction in the two-agent team, especially in identifying hidden error or result abnormality when no error/complain occurs during the executing of the simulation script. It is noted that since we can flexibly assign agents, future work could explore how fine-tuned agents or agents with special capabilities (e.g., writing or executing particular code packages via function calling, searching literature, databases, etc.) can further improve results.

## 3. Conclusions

Language is a critical concept not only for humans, generally, but specifically in scientific and engineering contexts, forming key ingredients in perceptions of the world, communication, teaching and others, like human-to-human interactions and collaborations, model building (e.g., mathematical language and coding, simulation engines), and many others. While still at their infancy, language-focused deep learning models such as LLMs present unprecedented capabilities in understanding and applying diverse collective knowledge in terms of various modalities[66]. When adapted to domain knowledge by further training or fine-tuning[46–48] they can gain a more specialized set of knowledge and skills. As shown here, the possibility to dynamically allocate, retrieve or generate new data, physical or empirical ground truths, and other pertinent information through the use of multi-agent interactions has the potential to enhance the current use of machine learning in science. For example, such strategies allow for numerical modeling methods that have been developed for mechanics or other engineering applications – which have strong physics-based foundations and can serve as the toolbox for generating large amount of domain-specific data. By incorporating these into physics-inspired multi-agent AI systems they thereby expand the concept of a "trained model" that relies solely on baked-in knowledge towards an active system that can retrieve physics-based data on demand, logically operate in a broader context of scientific problem solving, and educate itself to develop additional domain knowledge, collecting new data, optimizing the response, and using the insights garnered in the process to solve problems.

To synergize the strength of scientific AI and machine learning with pertinent advantages in physics-based numerical modeling, in this study, we have explored the potential of organizing sophisticated general-purpose LLM based agents to solve mechanics problems using numerical modeling tools to in an autonomous manner. The general concept of multi-agent AI systems is not limited to using LLMs as agents; they can possibly include a variety of additional special-purpose modeling and simulation tools, experimental capabilities for data collection (e.g., automated robotic systems), human input, and expert AI systems or surrogate models trained to solve particular tasks. Since the agent system can incorporate a variety of measures to assess the quality of generations or predictions, it can easily incorporate physical constraints (e.g., equilibrium conditions, mass balance, or check for soundness of solutions that may include manufacturability). With the emergence of fine-tuned or retrieval augmented systems, such a framework can learn further from experiences, human input, expert knowledge and other feedback, making them effective problem solvers. It is further noted that the general capabilities of the AI agents used are critical; in our case, the GPT-4 was able to properly construct codes and input files for complex FEM simulation scripts. If codes are used that are more complicated, additional fine-tuning (e.g., on code bases or manuals/documentation) may be necessary. Here, retrieval-augmented methods[67] can also play an important role. Further, the data generated – be it via a FEM simulation as done here, another simulation tool (er e.g. Density Functional Theory or molecular dynamics), or experiment – can be incorporated into the memory of the model (e.g. via prompt history, or by construction vector embedding databases). If these results (and mechanisms by which the results were produced) are shared with others via an API, for instance, other AI systems can access the data, input files used to produce the data, and so on, and thereby further accelerate progress. This emerging potential is an exciting frontier in scientific AI that has many potential applications across all areas of science, engineering, including mechanics, which requires new models for information and data sharing[68]. Building on the work done here, agent-based models could also be used in future research to produce their own surrogate models using various ML techniques – e.g., we can instruct models train a model against a set of generated data and then utilize these newly emerging tools in their problem-solving strategy. What's more, the use of distinct agents allows us to build in various checks and balances – e.g., we can define an agent that checks the results against physical principles (e.g., does the solution field satisfy equilibrium conditions, or a certain constraint in a design problem, etc.). There is also no intrinsic limitation in terms of what the AI agents can do; e.g., here we used a set of LLMs to solve problems, but we can also define agents that focus on a particular area simulation (e.g., an expert in running DFT, MD, physics-inspired neural network solvers[69–71], etc.). This integrative potentiality is already implemented within the OpenAI ecosystem and some open-source tools via function calling; and such agent-based AI systems can be particularly appealing as it allows us to take advantage of diverse sets of tools developed by the community (and for which the LLM may not be able to write code on its own). The coding ability in various languages further provides very broad capabilities to achieve generation of new knowledge, and organizing it along with diverse feedback obtained by the agent team, for the purpose of problem enhanced solving.

To go a bit deeper, as shown in this Letter, the agents powered by a general-purpose large language model, such as GPT-4, offer comprehensive knowledge in executing classical mechanics theories and well-developed numerical methods, such as FEM. While such models do not serve as proxies to provide the solution to problems, they excel at developing a strategy to solve problems, e.g. via numerical computation or in future work, empirical data collection by designing experiments or even research hypothesis as part of larger data collection frameworks[72]. We have focused on this perspective by unleashing those capabilities in a series of computational experiments that feature different organizational structures by which a multi-agent modeling framework can be constructed. With a basic two-agent team, we have demonstrated that via continuous conversation, the agents can make self-corrections based on the outcomes of the modeling results continuously. This feature enables the team to robustly apply FEM and autonomously solve classical elasticity problems of different variants, including different boundary conditions, domain geometries and meshes, small/finite deformation and linear/hyper-elastic constitutive laws. For complex tasks where self-correction suffers from lacking insightful inputs or obvious coding errors, we have constructed a multi-agent group using the idea of division of labor. To do so, we assign specialized functional roles to agents and organize them dynamically in a group chat. By profiling through initial prompts, we assign different focuses to the agents, including plan making, problem formulating, code implementing and debugging, simulation executing as well as performance criticizing. During an AI group chat, we use a chat manager agent to dynamically choose speakers based on the current discussion outcome and the agents' roles, thus enabling an adaptive topology of the group chat. With such a dynamic multi-agent group, we have demonstrated that individual agents perform better with clear focuses and benefit from mutual corrections emerged during group chatting. For example, the agent playing the role of a critic is able to identify abnormality in simulation results even when there is no coding error reported and provide key improvements in completing the whole task. These types of mutual corrections go beyond self-corrections and thus, the multi-agent group is able to solve more challenging tasks that the two-agent team fail in correctly retrieving and identifying the problem type, retrieving relevant material properties and handling complex postprocessing.

Our case studies on the multi-agent modeling frameworks have demonstrated great potentials in amplifying the capability of conservable agents via suitable organizations as well as integrating AI-agents into physics-based modeling for automation, thus preparing for a human-AI teaming up future for solving various engineering and scientific problems. From one perspective, the self-correction capacity and mutual corrections observed in the two-agent and multi-agent teams strongly indicate that the interaction between agents plays key roles in affecting their overall performances as they achieve certain synergies that single AI agents could not achieve on their own. Future studies can focus on improving/optimizing the role design and chatting topology in the multi-agent team for solving specific engineering problems. On another front, the automation of mechanics modeling through multi-agent frameworks may enhance the efficiency of applying modeling tools for many other engineering or science applications, including a variety of other modeling techniques such as first-principles calculations, molecular modeling and multiscale integration. By achieving robust handling of different variants and code errors, future study can take advantage of this self-piloting conversable modeling framework to curate large datasets with rich domain knowledge, discover novel structure/material designs for superior mechanical properties as well as teaching mechanics theory and modeling methods in a human-AI interactive manner. This is especially promising when computational methods can be directly coupled with experimental platforms that offer high-throughput data retrieval[73–75].

## 4. Materials and Methods

**Agent design and setup**

As shown in **Figure 1a**, we construct and organize AI agents using GPT-4[40] and the AutoGen framework[76], an open-source platform to implement agent-based AI modeling.[63]

The AutoGen framework provides the structural backbone for the orchestration, optimization, and automation of workflows by which the individual AI agents interact. AutoGen supports the development of applications using multiple conversable agents that can interact with each other; they are highly customizable and can work in different modes that incorporate a mix of LLMs, code development and execution, human input, and various tools that can be programmed via function calling (in this case, the agent has a special capability to call a set of predefined functions which then performs an operation, such as executing a Python code, conducting a web search, solving a DFT problem, generating an image, and returns the result back to the agent). Here, we do not use function calling but instead use the LLM's innate capability to write code.

As LLM we use the model "gpt-4-0613" with a context window of 8,192 tokens and training data up to September 2021 from OpenAI to power the agents via OpenAI API ports.[64] AutoGen can also work with Open Source LLMs, in principle, albeit this is not explored here.

In the two-agent team, the assistant agent is constructed using the AssistantAgent class from AutoGen and the human user proxy agent is created using UserProxyAgent class from AutoGen. In the multi-agent group, the admin and executor are created using UserProxyAgent class; the planner, scientist, engineer and critic are created using AssistantAgent class; and the group chat manager is created using GroupChatManager class. The role profiles for each of the agents listed in **Table 4** are included as system message at their creation. We use the python package, FEniCS[65] as the simulation environment for solving elasticity problems using FEM. The codes in the GitHub repository include all other necessary details used in the setup.

**Software versions and hardware**

We develop our multi-agent models using Google Colab.[77,78] We use Python 3.10,[79] pyautogen-0.1.14[76] and install FEniCS via the FEM-on-Colab package. [80,81]

**Acknowledgments:** We acknowledge support from USDA (2021-69012-35978), DOE-SERDP (WP22-S1-3475), ARO (79058LSCSB, W911NF-22-2-0213 and W911NF2120130) as well as the MIT-IBM Watson AI Lab and MIT's Generative AI Initiative. Additional support from NIH (U01EB014976 and R01AR077793) and ONR (N00014-19-1-2375 and N00014-20-1-2189) is acknowledged.

**Conflict of interest**

The author declares no conflict of interest.

**Data and code availability**

Data and codes are either available on GitHub at https://github.com/lamm-mit/MechAgents or will be provided by the corresponding author based on reasonable request.

**Author contributions:** MJB and BN conceived the study. BN developed the multi-agent models, performed the tests for various problems, analyzed the results and prepared the first draft of the paper. MJB supported the analysis, revised and finalized the paper with BN.

**Supplementary Materials**

The full records of different conversation tests and the result files generated are provided as **Supplementary Materials**.

*S1. Two-agent conversation on solving elasticity problem using finite element method*

    S1.1 Conversation 1: solving a linear elasticity problem

    Tables S1-S5

    S1.2 Conversation 2: solving a hyperelasticity problem

    Tables S6-S7

    S1.3 Conversation 3: A more challenging task

    Table S8

*S2. Multi-agent conversation on solving elasticity problem using finite element method*

    S2.1 Group chat 1: Matching the performance of the two-agent model

    Tables S9-S12

    S2.2 Group chat 2: Solving the challenging task in Conversation S1.3

    Table S13

**Figures and Tables**

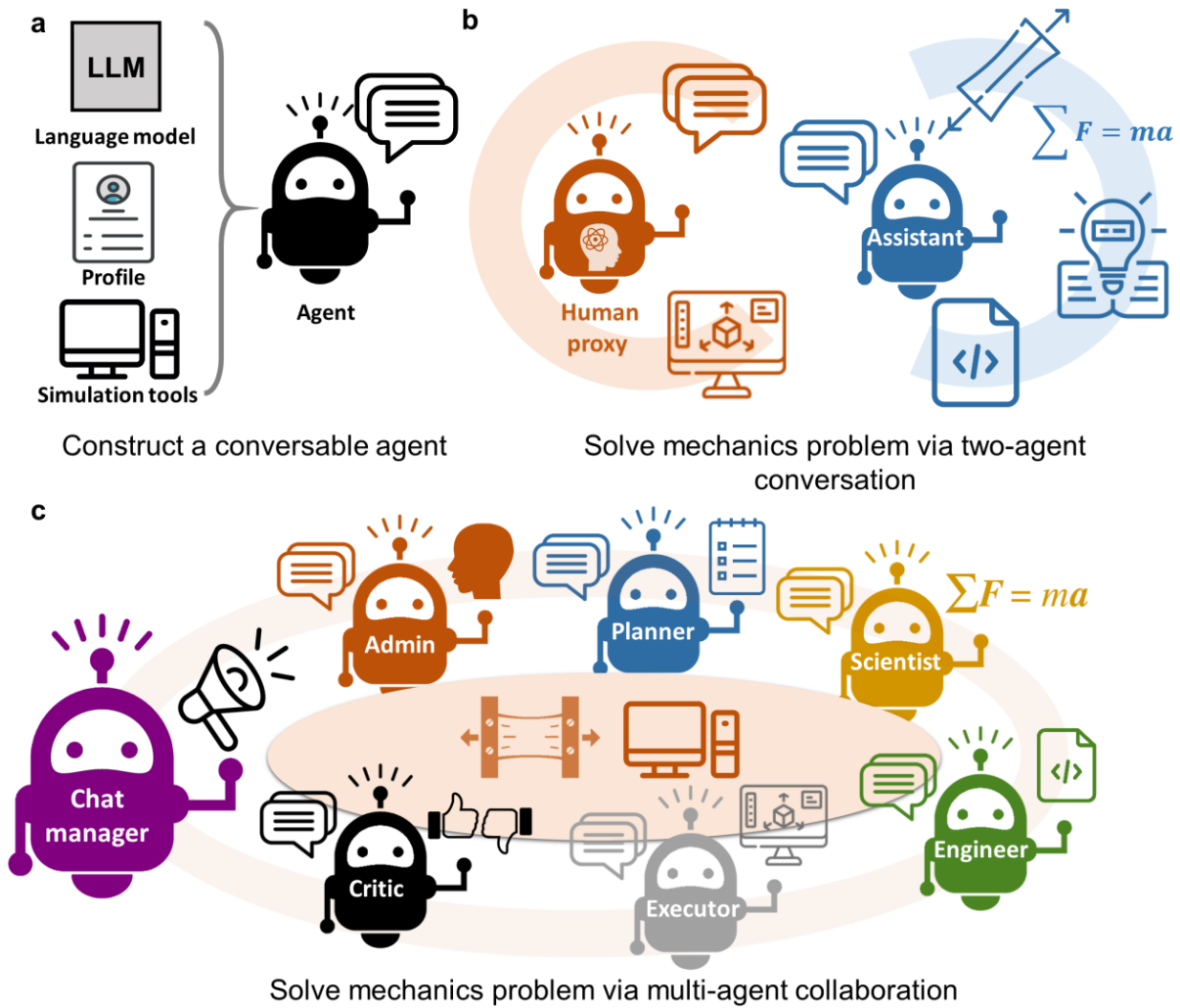

**Figure 1:** Framework of multi-agent AI modeling for solving mechanics problems. **a**, Constructing a conversable agent that can talk, has a focus and use simulation tools. **b**, The structure of a two-agent team, where one agent serves as the human proxy for giving the assignment, running simulations code and returning outcomes while the other agent is in charge of formulating a plan, retrieving relevant knowledge, writing the code to solve the problem, analyzing and debugging the outcome of the code by self-correcting. **c**, The structure of a multi-agent team that allows for a division of labor and mutual correction via dynamical, autonomous group chat. The working agents are given different profiles to focusing on plan making, problem formulating, code writing/debugging, tool executing and result criticizing. Via collaboration under the leading chat manager agent, they can solve more challenging tasks that exceed the capability of a two-agent team.

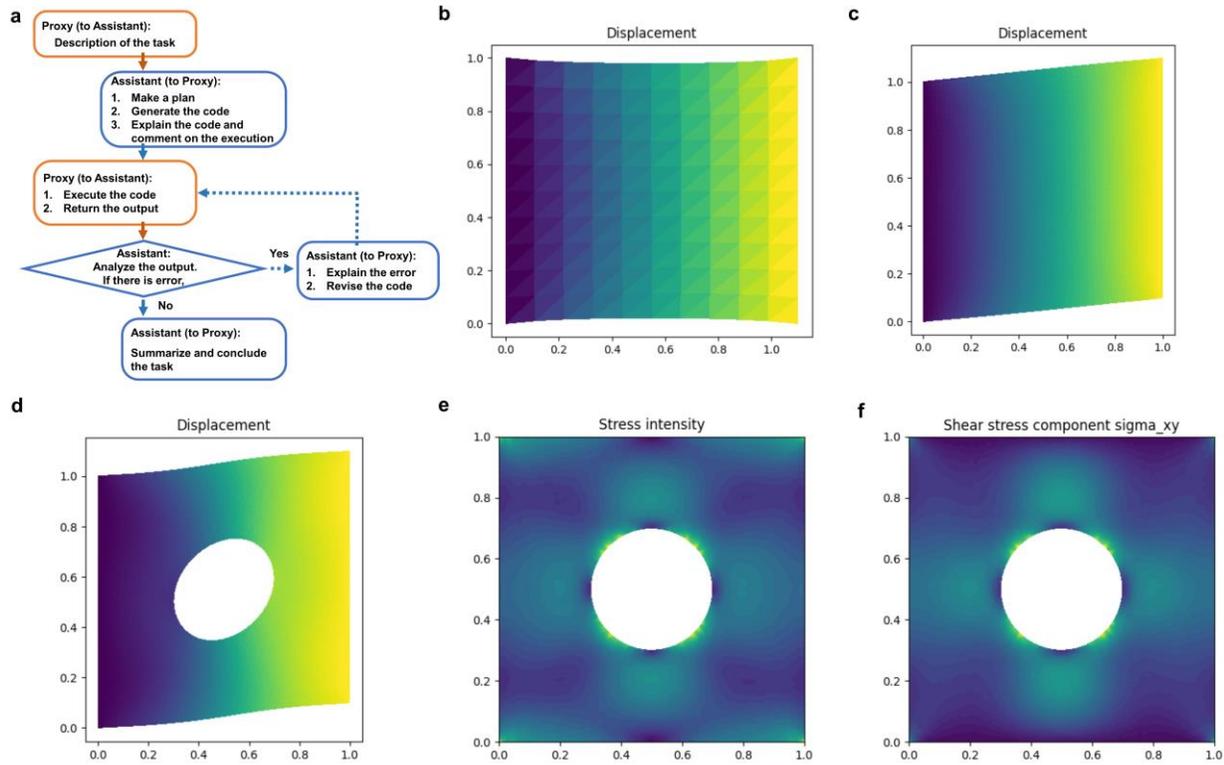

**Figure 2:** Conversation flow observed, and result files generated by the two-agent team in Conversation 1. **a**, Summary of a typical conversation flow observed in one round of conversation between the two agents, during which the assistant agent often can correct itself (dash arrows) based on the outcome provided by the proxy agent. **b-d**, Predicted displacement fields for Rounds 1-3 in Conversation 1. **e-f**, Predicted stress fields as identified in Rounds 4-5.

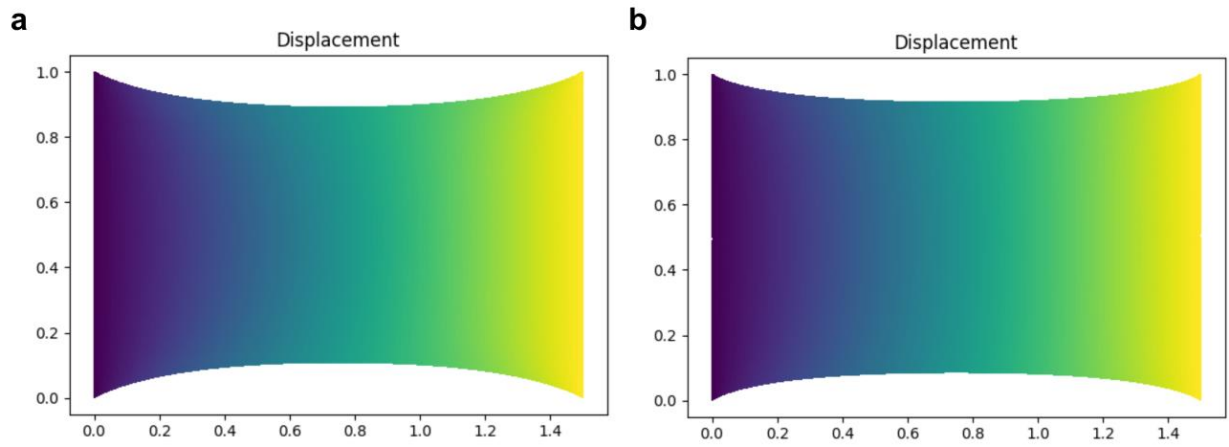

**Figure 3:** Results generated during Conversation 2 by the two-agent team. **a**, Predicted displacement field result using a linear elasticity model with small deformation assumption. **b**, Displacement field result using a hyperelastic Neo-Hookean model considering finite deformation.

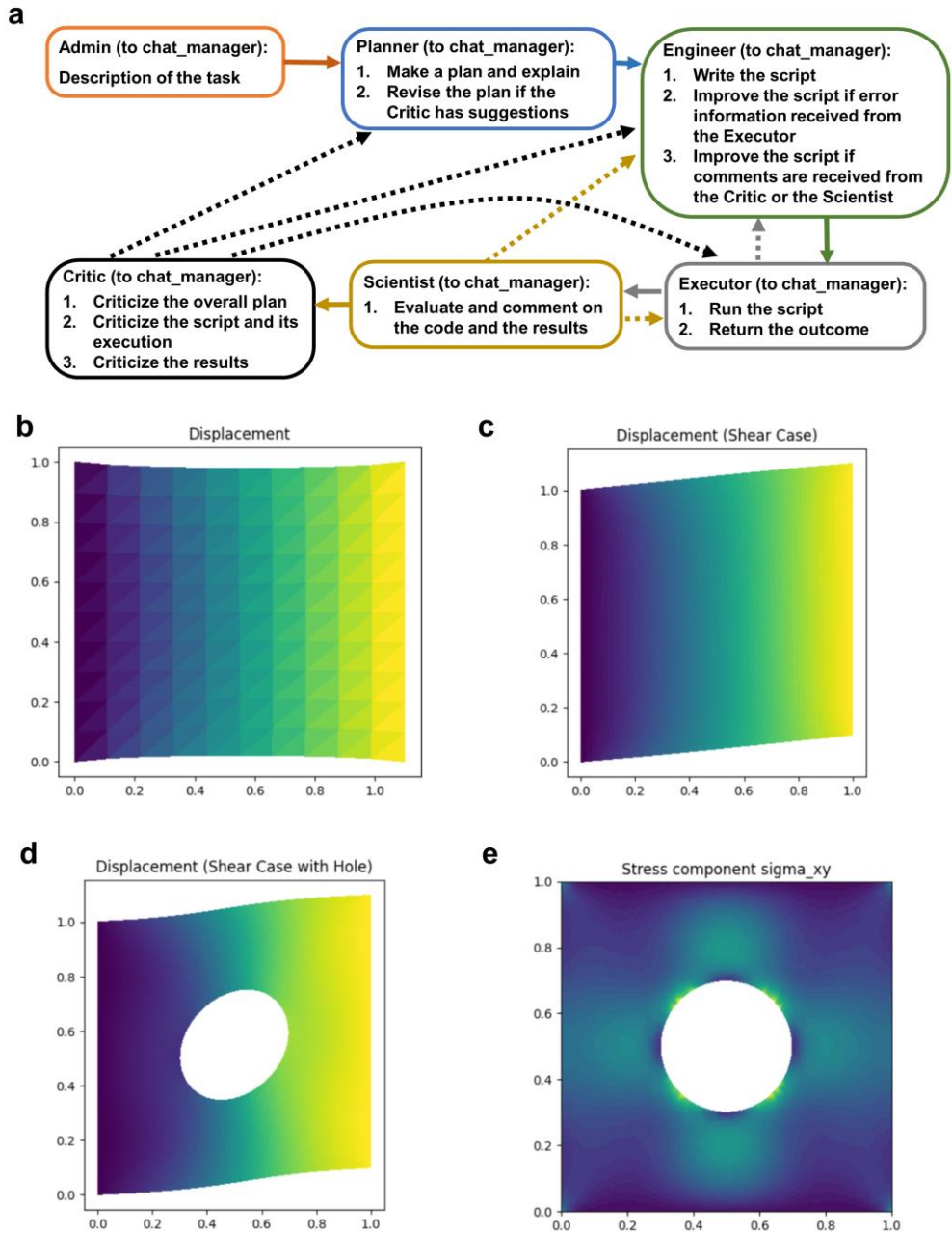

**Figure 4:** Dynamic interactions between agents in the group chat and the result plots generated for linear elastic problems. **a**, Dynamically organized by the group chat manager agent, the working agents are selected to speak or response based on the current context of the chat, thus forming close interactions and achieving mutual corrections. **b-e**, Result files generated directly by the group chat, without human intervention, for the assignments listed in Row 2-4 Column 2 in **Table 1.**

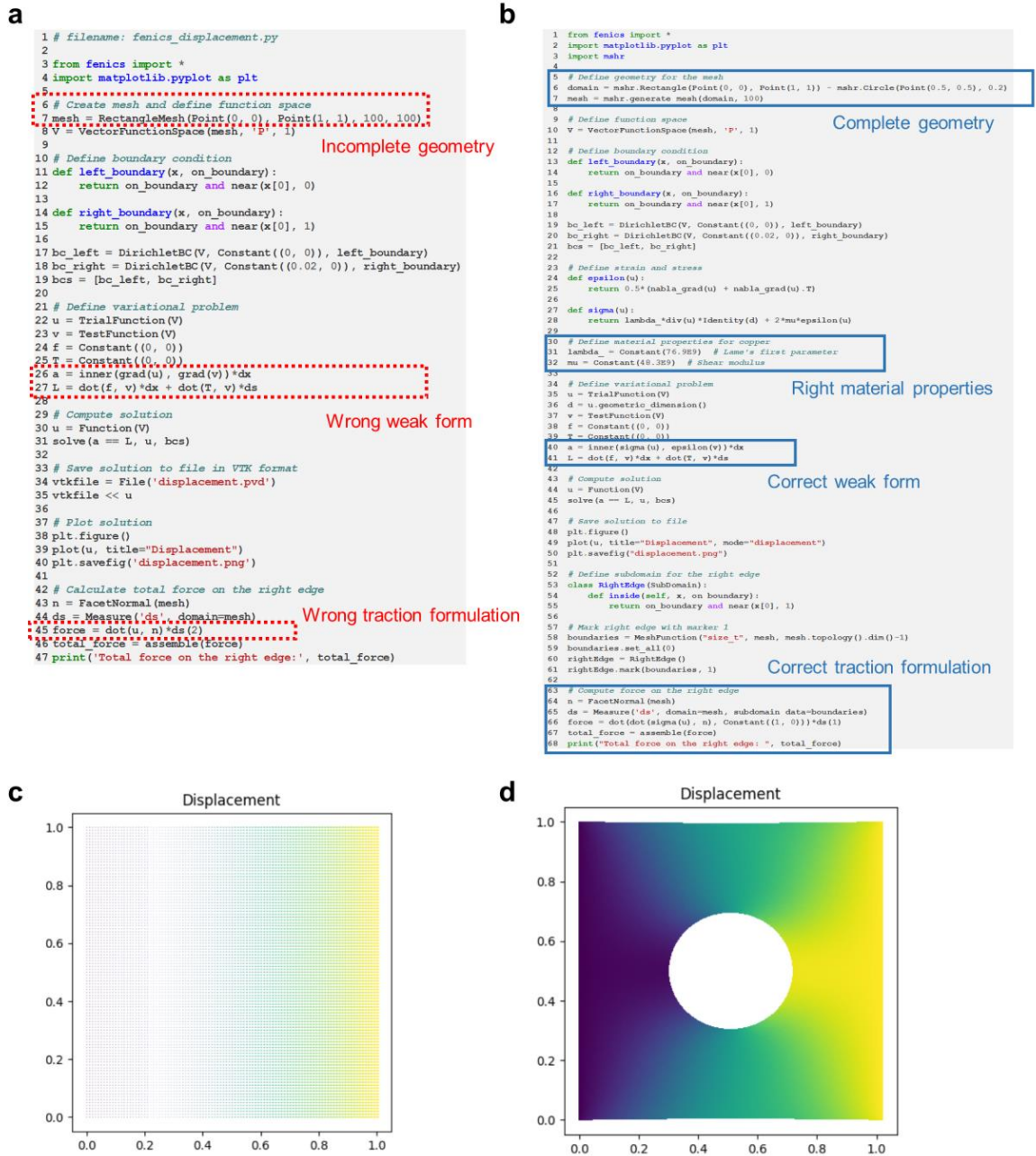

**Figure 5:** The final scripts and results generated by the two-agent team and the multi-agent group with division of labor for solving the uniaxial tension assignment in Group chat 2. **a**, The FEniCS script by the two-agent team defines an incomplete geometry, formulates wrong weak form for linear elasticity problem and uses wrong formula to calculate traction force on the right edge (highlighted in the red dash rectangles). **b**, The script by the multi-agent group is correct after multiple rounds of revision via group chatting and mutual correction. **c**, The result file generate by the two-agent team (c), which is totally wrong. **d**, The result file of displacement filed generated by the multi-agent group, which is correct.

**Table 1**: Summary of Conversation 1 by the two-agent team in solving linear elasticity problems using a finite element method. Tasks (Column 2) from human user are assigned through the proxy agent at the beginning of the multiple rounds (Column 1) during the conversation, starting with a clearly stated uniaxial tension problem in linear elasticity, followed by variants with different boundary conditions, FEM meshes, domain geometries, postprocessing for stress components. The responses by the assistant agent are summarized in Column 3 (highlighted in blue for success and in red for failures) and commented in Column 4. Note that, within each round, no human input is provided in the middle. The two-agent team successfully handles the first three rounds of tasks via self-correction during conversations, fails at Round 4 without self-realization of the error, and then correct its mistake at Round 5 with the hint from the new assignment (Column 2 Row 6).

| Round # | Assignments from the proxy agent to the assistant agent | Summary of tasks performed by the assistant agent | Notes |
|---|---|---|---|
| 1 | A 1m-by-1m elastic plate has Young's modulus of 1GPa and Poisson ration of 0.3. It has zero displacement on the left edge and 0.1m displacement along x direction on the right edge. Solve for the displacement using FENICS and store the displacement result in a PNG file. | Write the code; Explain the plan; Analyze the error information of the code and revise the code twice; Conclude the task. | Succeed; Can formulate linear elasticity problem using FEM; can handle coding errors in defining material properties and stress tensor. |
| 2 | Let's change the boundary condition on the right edge to a shear case, i.e., the displacement along y direction is 0.1m on the right edge. Please refine the mesh to 50-by-50 elements, solve the problem again and save result into another png file. | Revise the code for boundary condition and mesh parts; Explain the plan; Conclude the task. | Succeed; Can implement right boundary condition and FEM mesh |
| 3 | Let's add a circular hole of radius 0.2m in the middle of the original square domain. Please solve the share problem and save result into another png file. | Revise the code for domain definition part; Explain the modification; Conclude the task. | Succeed; Can implement correct geometry |
| 4 | Let's also calculate the stress component sigma_xy and save it into another png file. | Revise the code for stress post-processing and save von Mises stress; Analyze the error information of the code and revise the code twice; Conclude the task. | Fail by mistaking von Mises stress for the required shear stress; The final script is error-free; Fail to realize its failure. |
| 5 | You plotted von Mises stress, not stress component sigma_xy. Please correct | Revise the code by extracting the shear stress component; Explain the modification; Conclude the task. | Succeed in identifying the right stress component. |

**Table 2:** Summary of Conversation 2 by the two-agent team in solving the linear/hyper-elasticity problems using finite element method. The format of the table is the same with **Table 1**.

| Round # | Assignments from the proxy agent to the assistant agent | Summary of tasks performed by the assistant agent | Notes |
|---|---|---|---|
| 1 | A 2D plate occupies 1m-by-1m domain. It is assumed as linear elastic and has Young's modulus of 1GPa and Poisson ration of 0.3. It has zero displacement on the left edge and 0.5m displacement along x direction on the right edge. The top and bottom edges are free to move. Solve for the displacement using FENICS with a fine mesh of 100by100 elements and store the displacement result in a PNG file named 1.png. | Write the code; Explain the plan; Analyze the error information of the code and revise the code twice; Conclude the task. | Succeed; Can formulate linear elasticity problem using FEM; Can handle coding errors in defining material properties and stress tensor. |
| 2 | Let's change the problem from linear elasticity into hyperelasticity. Please use a compressible Neo-hookean model with Young's modulus of 1GPa and Poisson ration of 0.3, please consider finite deformation, resolve the nonlinear problem and save result into another png file named 2.png. | Revise the code; Explain the plan; Analyze the error information of the code and revise the code once; Conclude the task. | Succeed; Can formulate hyperelasticity problem with finite strain using FEM; Can handle coding errors in defining nonlinear problem in FEniCS. |

**Table 3:** FEniCS scripts generated by the two-agent AI team after self-corrections, for uniaxial tension problems with linear elastic (left column) and hyperelastic (right column) materials.

| Linear elasticity case | Hyperelasticity case |
|---|---|
| ```python
# filename: fenics_script.py

from fenics import *
import matplotlib.pyplot as plt

# Create mesh and define function space
mesh = UnitSquareMesh(100, 100)
V = VectorFunctionSpace(mesh, 'P', 1)

# Define boundary condition
tol = 1E-14

def left_boundary(x, on_boundary):
    return on_boundary and near(x[0], 0, tol)

def right_boundary(x, on_boundary):
    return on_boundary and near(x[0], 1, tol)

bc_left = DirichletBC(V, Constant((0, 0)), left_boundary)
bc_right = DirichletBC(V, Constant((0.5, 0)), right_boundary)
bcs = [bc_left, bc_right]

# Define strain and stress

def epsilon(u):
    return 0.5*(nabla_grad(u) + nabla_grad(u).T)

def sigma(u):
    E = 1e9   # Young's modulus
    nu = 0.3  # Poisson's ratio
    mu = E / (2*(1 + nu))
    lmbda = E*nu / ((1 + nu)*(1 - 2*nu))
    return lmbda*div(u)*Identity(d) + 2*mu*epsilon(u)

# Define variational problem
u = TrialFunction(V)
d = u.geometric_dimension()  # space dimension
v = TestFunction(V)
f = Constant((0, 0))
T = Constant((0, 0))
a = inner(sigma(u), epsilon(v))*dx
L = dot(f, v)*dx + dot(T, v)*ds

# Compute solution
u = Function(V)
solve(a == L, u, bcs)

# Save solution to file in VTK format
vtkfile = File('solution.pvd')
vtkfile << u

# Plot solution
plt.figure()
plot(u, title='Displacement', mode='displacement')

# Save plot to file
plt.savefig('1.png')
``` | ```python
# filename: fenics_script.py

from fenics import *
import matplotlib.pyplot as plt

# Create mesh and define function space
mesh = UnitSquareMesh(100, 100)
V = VectorFunctionSpace(mesh, 'P', 1)

# Define boundary condition
tol = 1E-14

def left_boundary(x, on_boundary):
    return on_boundary and near(x[0], 0, tol)

def right_boundary(x, on_boundary):
    return on_boundary and near(x[0], 1, tol)

bc_left = DirichletBC(V, Constant((0, 0)), left_boundary)
bc_right = DirichletBC(V, Constant((0.5, 0)), right_boundary)
bcs = [bc_left, bc_right]

# Define strain and stress
E = 1e9   # Young's modulus
nu = 0.3  # Poisson's ratio
mu = E / (2*(1 + nu))
lmbda = E*nu / ((1 + nu)*(1 - 2*nu))

def psi(F):
    I = Identity(F.geometric_dimension())  # Identity tensor
    C = F.T*F  # Right Cauchy-Green tensor
    Ic = tr(C)
    J = det(F)
    return (mu/2)*(Ic - 2) - mu*ln(J) + (lmbda/2)*(ln(J))**2

# Define variational problem
u = Function(V)
d = u.geometric_dimension()  # space dimension
I = Identity(d)
F = I + grad(u)  # Deformation gradient
Pi = psi(F)*dx
F = derivative(Pi, u, TestFunction(V))

# Compute solution
solve(F == 0, u, bcs, solver_parameters={"newton_solver":{"rel

# Save solution to file in VTK format
vtkfile = File('solution.pvd')
vtkfile << u

# Plot solution
plt.figure()
plot(u, title='Displacement', mode='displacement')

# Save plot to file
plt.savefig('2.png')
``` |

**Table 4:** The profiles of the agents used in the multi-agent group setup following the concept of division of labor.

| Agent # | Agent role | Agent profile |
|---|---|---|
| 1 | Human proxy Admin | A human admin. Interact with the planner to discuss the plan. Plan execution needs to be approved by this admin. |
| 2 | Planner | Planner. Suggest a plan. Revise the plan based on feedback from admin and critic, until admin approval. <br> The plan may involve an engineer who can write code and a scientist who doesn't write code. <br> Explain the plan first. Be clear which step is performed by an engineer, and which step is performed by a scientist |
| 3 | Scientist | Scientist. You follow an approved plan. <br> You are able to figure out 1. geometry of the mesh; 2. boundary conditions; 3. constitutive law of materials and related material properties and 4. formulate the mechanics problem. <br> You don't write or execute code. <br> You explicit check the boundary results with the given boundary conditions. |
| 4 | Engineer | Engineer. You follow an approved plan. <br> You write python code to solve tasks using FENICS. <br> You pay attention to mesh, boundary conditions, material properties and constitutive law. <br> Wrap the code in a code block that specifies the script type. The user can't modify your code. So do not suggest incomplete code which requires others to modify. Don't use a code block if it's not intended to be executed by the executor. <br> Don't include multiple code blocks in one response. Do not ask others to copy and paste the result. Check the execution result returned by the executor. <br> If the result indicates there is an error, fix the error and output the code again. Suggest the full code instead of partial code or code changes. If the error can't be fixed or if the task is not solved even after the code is executed successfully, analyze the problem, revisit your assumption, collect additional info you need, and think of a different approach to try. <br> When writing code, assert the boundary condition. <br> You don't install packages. |
| 5 | Executor | No special profile needed; focuses on executing the code and return the outcomes. |
| 6 | Critic | Critic. <br> You double check plan, claims, code from other agents, results on the boundary conditions and provide feedback. <br> Check whether the plan includes adding verifiable info such as satisfying boundary conditions, having source URL. |
| 7 | Group chat manager | This agent repeats the following steps: Dynamically selecting a speaker, collecting response, and broadcasting the message to the group. |